\documentclass[11pt,letterpaper]{article}
\usepackage{emnlp2017}
\usepackage{times}
\usepackage{latexsym}
\usepackage{amssymb}
\usepackage{amsmath}
\usepackage{algcompatible}
\usepackage{algorithm}
\usepackage{graphicx}
\usepackage{url}
\usepackage{bm}
\usepackage{color}
\usepackage{booktabs}
\usepackage{afterpage}

\newcommand{\vect}[1]{\bm{#1}}
\newcommand{\matr}[1]{\bm{#1}}

\newcommand{\va}[0]{\vect{a}}
\newcommand{\vb}[0]{\vect{b}}
\newcommand{\vc}[0]{\vect{c}}
\newcommand{\ve}[0]{\vect{e}}
\newcommand{\valpha}[0]{\vect{\alpha}}
\newcommand{\vgamma}[0]{\vect{\gamma}}
\newcommand{\vbeta}[0]{\vect{\beta}}

\newcommand{\vh}[0]{\vect{h}}
\newcommand{\vv}[0]{\vect{v}}

\newcommand{\vz}[0]{\vect{z}}

\newcommand{\vs}[0]{\vect{s}}

\newcommand{\vi}[0]{\vect{i}}
\newcommand{\vo}[0]{\vect{o}}

\newcommand{\vr}[0]{\vect{r}}

\newcommand{\mW}[0]{\matr{W}}

\newcommand{\mE}[0]{\matr{E}}

\newcommand{\mU}[0]{\matr{U}}

\newcommand{\mL}{\matr{L}}

\emnlpfinalcopy

\def\emnlppaperid{1026}

\title{An empirical study on the effectiveness of images in Multimodal Neural Machine Translation}

 \author{Jean-Benoit Delbrouck \and St\'ephane Dupont \\
         TCTS Lab, University of Mons, Belgium\\
          \{jean-benoit.delbrouck, stephane.dupont\}@umons.ac.be}

\date{}

\begin{document}
	
	\maketitle
	
	\begin{abstract}
		In state-of-the-art Neural Machine Translation (NMT), an attention mechanism is used during decoding to enhance the translation. At every step, the decoder uses this mechanism to focus on different parts of the source sentence to gather the most useful information before outputting its target word. Recently, the effectiveness of the attention mechanism has also been explored for multimodal tasks, where it becomes possible to focus both on sentence parts and image regions that they describe. In this paper, we compare several attention mechanism on the multimodal translation task (English, image $\rightarrow$ German) and evaluate the ability of the model to make use of images to improve translation. We surpass state-of-the-art scores on the Multi30k data set, we nevertheless identify and report different misbehavior of the machine while translating. 
	\end{abstract}
	
	\section{Introduction}
	
	In machine translation, neural networks have attracted a lot of research attention. Recently, the attention-based encoder-decoder framework \cite{SutskeverVL14,BahdanauCB14} has been largely adopted. In this approach, Recurrent Neural Networks (RNNs) map source sequences of words to target sequences. The attention mechanism is learned to focus on different parts of the input sentence while decoding. Attention mechanisms have shown to work with other modalities too, like images, where their are able to learn to attend the salient parts of an image, for instance when generating text captions \cite{icml2015_xuc15}. For such  applications, Convolutional Neural Networks (CNNs) such as Deep Residual \cite{He_2016_CVPR} have shown to work best to represent images.
	
	Multimodal models of texts and images empower new applications such as visual question answering or multimodal caption translation. Also, the grounding of multiple modalities against each other may enable the model to have a better understanding of each modality individually, such as in natural language understanding applications. 
	
	In the field of Machine Translation (MT), the efficient integration of multimodal information still remains a challenging task. It requires combining diverse modality vector representations with each other. These vector representations, also called context vectors, are computed in order the capture the most relevant information in a modality to output the best translation of a sentence.  
	
	To investigate the effectiveness of information obtained from images, a multimodal machine translation shared task \cite{specia-EtAl:2016:WMT} has been  addressed to the MT community\footnote{http://www.statmt.org/wmt16/multimodal-task.html}. The best results of NMT model were those of \citeauthor{huang2016attention} \shortcite{huang2016attention} who used LSTM fed with global visual features or multiple regional visual features followed by rescoring. Recently, \citeauthor{CalixtoLC17b} \shortcite{CalixtoLC17b} proposed a doubly-attentive decoder that outperformed this baseline with less data and without rescoring. 
	
	Our paper is structured as follows. In section \ref{nmt}, we briefly describe our NMT model as well as the conditional GRU activation used in the decoder. We also explain how multi-modalities can be implemented within this framework. In the following sections (\ref{ABM} and \ref{iaopti}), we detail three attention mechanisms and explain how we tweak them to work as well as possible with images. Finally, we report and analyze our results in section \ref{experiment} then conclude in section \ref{concl}.
	
	\section{Neural Machine Translation}\label{nmt}
	
	In this section, we detail the neural machine translation architecture by \citeauthor{BahdanauCB14} \shortcite{BahdanauCB14}, implemented as an attention-based encoder-decoder framework with recurrent neural networks (\S\ref{textbasednmt}). We follow by explaining the conditional GRU layer (\S\ref{cgrusec}) - the gating mechanism we chose for our RNN - and how the model can be ported to a multimodal version (\S\ref{multinmt}).
	
	\subsection{Text-based NMT}\label{textbasednmt}
	Given a source sentence $X = (x_1, x_2, \hdots , x_M)$, the neural network directly models the conditional probability $p(Y|X)$ of its translation $Y = (y_1, y_2, \hdots, y_N)$. The network consists of one encoder and one decoder with one attention mechanism. The encoder computes a representation $C$ for each source sentence and a decoder
	generates one target word at a time and by decomposing
	the following conditional probability :
	\begin{equation}\log p(Y|X) = \sum\limits_{t=1}^n \log p(y_t | y<t, C)\end{equation}
	Each source word $x_i$ and target word $y_i$ are a column index of the embedding matrix $\mE_X$ and $\mE_Y$. The encoder is a bi-directional RNN with Gated Recurrent Unit (GRU) layers \cite{ChungGCB14,cho-al-emnlp14}, where a forward RNN $\overrightarrow{\Psi}_\text{enc}$ reads the input sequence as it is ordered (from $x_1$ to $x_M$) and calculates a sequence of forward hidden states $(\overrightarrow{\vh}_1, \overrightarrow{\vh}_2, \hdots, \overrightarrow{\vh}_M)$. A backward RNN $\overleftarrow{\Psi}_\text{enc}$ reads the sequence in the reverse order (from $x_M$ to $x_1$), resulting in a sequence of backward hidden states $(\overleftarrow{\vh}_M, \overleftarrow{\vh}_{M-1}, \hdots, \overleftarrow{\vh}_1)$. We obtain an annotation for each word $x_i$ by concatenating the forward and backward hidden state $\vh_t = [\overrightarrow{\vh}_t;\overleftarrow{\vh}_t]$. Each annotation $\vh_t$ contains the summaries of both the preceding words and the following words. The representation $C$ for each source sentence is the sequence of annotations $C = (\vh_1, \vh_2, \hdots, \vh_M)$.\\ \\
	The decoder is an RNN that uses a conditional GRU (cGRU, more details in \S\ref{cgrusec}) with an attention mechanism to generate a word $y_t$ at each time-step $t$. The cGRU uses it's previous hidden state $\vs_{t-1}$, the whole sequence of source annotations $C$ and 
	the previously decoded symbol $y_{t-1}$ in order to update it's hidden state $\vs_t$ : 
	\begin{equation}
	\mathbf{s}_t = \text{cGRU}\left(  \mathbf{s}_{t-1}, y_{t-1}, C  \right)
	\end{equation}
	In the process, the cGRU also computes a time-dependent context vector $\vc_t$. Both $\vs_{t}$ and $\vc_t$ are further used to decode the next symbol. We use a deep output layer \cite{Pascanu2014} to compute a vocabulary-sized vector :
	\begin{equation}
	\vo_t=\mL_o \tanh (\mL_ss_t + \mL_c\vc_t + \mL_w\mE_Y[y_{t-1}]) \label{deepout}
	\end{equation}
	where $\mL_o$, $\mL_s$, $\mL_c$, $\mL_w$ are model parameters. We can parameterize the probability of decoding each word $y_t$ as:
	\begin{equation}
	p(y_t | y_{t-1}, \vs_t, \vc_t) = \text{Softmax}(\vo_t)
	\end{equation}
	The initial state of the decoder $\vs_0$ at time-step $t=0$ is initialized by the following equation : 
	\begin{equation}
	\vs_0 = f_{\text{init}}(\vh_M) \label{initdec}
	\end{equation}
	where $f_{\text{init}}$ is a feedforward network with one hidden layer.
	
	\subsection{Conditional GRU} \label{cgrusec}
	The conditional GRU \footnote{\url{https://github.com/nyu-dl/dl4mt-tutorial/blob/master/docs/cgru.pdf}} consists of two stacked GRU activations called $\text{REC}_1$ and $\text{REC}_2$ and an attention mechanism $f_{\text{att}}$ in between (called ATT in the footnote paper).	At each time-step $t$, REC1 firstly computes a hidden state proposal $\vs_t$ based on the previous hidden state $\vs_{t-1}$ and
	the previously emitted word $y_{t-1}$:
	\begin{align}
	\vz_t^{\prime} =& ~ \sigma \left(  \mW_z^{\prime} \mE_Y[y_{t-1}] + \mU_z^{\prime} \vs_{t-1}  \right) \nonumber \\
	\vr_t^{\prime} =& ~ \sigma \left(  \mW_r^{\prime} \mE_Y[y_{t-1}] + \mU_r^{\prime} \vs_{t-1}  \right)  \nonumber \\ 
	\underline{\vs}_t^{\prime} =& ~\text{tanh} \left(   \mW^{\prime} \mE_Y[y_{t-1}] + \vr_t^{\prime} \odot (\mU^{\prime}\vs_{t-1})  \right)  \nonumber \\        
	\vs_t^{\prime} =& (1 - \vz_t^{\prime}) \odot \underline{\vs}_t^{\prime} + \vz_t^{\prime} \odot \vs_{t-1}
	\end{align}
	Then, the attention mechanism computes $c_t$ over the source sentence using the annotations sequence $C$ and the intermediate hidden state proposal $\vs_t^{\prime}$:
	\begin{equation}
	\vc_t = f_{\text{att}} \left(  \text{C}, \vs_t^{\prime}  \right)\label{fatt} 
	\end{equation}

    \cite{}
	Finally, the second recurrent cell $\text{REC}_2$, computes the hidden state $\vs_t$ of the $\text{cGRU}$ by looking at the intermediate representation $\vs_t^{\prime}$ and context vector $\vc_t$:
	\begin{align}
	\vz_t =& \sigma \left( \mW_z \vc_t + \mU_z \vs_t^{\prime} \right) \nonumber \\
	\vr_t =& \sigma \left( \mW_r \vc_t + \mU_r \vs_t^{\prime} \right) \nonumber \\
	\underline{\vs}_t =& \text{tanh} \left(  \mW \vc_t  + \vr_t \odot (\mU \vs_t^{\prime} )  \right)  \nonumber \\       
	\vs_t =& (1 - \vz_t) \odot \underline{\vs}_t + \vz_t \odot \vs_t^{\prime}
	\end{align}	
	\subsection{Multimodal NMT} \label{multinmt}
	Recently, \citeauthor{CalixtoLC17b} \shortcite{CalixtoLC17b} proposed a doubly attentive decoder (referred as the "MNMT" model in the author's paper) which can be seen as an expansion of the attention-based NMT model proposed in the previous section. Given a sequence of second a modality annotations $I=(\va_1,\va_2, \hdots, \va_L)$, we also compute a new context vector based on the same intermediate hidden state proposal $\vs_t^{\prime}$: 
	\begin{equation}
	\vi_t = f_{\text{att}}^\prime \left(  \text{I}, \vs_t^{\prime}  \right) 
	\end{equation}	
	This new time-dependent context vector is an additional input to a modified version of REC2 which now computes the
	final hidden state $\vs_t$ using the intermediate hidden state proposal $\vs_t^{\prime}$ and both time-dependent context vectors $\vc_t$ and $\vi_t$ :	
	\begin{align}
	\vz_t =& \sigma \left( \mW_z \vc_t + \mW_z \vi_t + \mU_z \vs_t^{\prime} \right)  \nonumber \\
	\vr_t =& \sigma \left( \mW_r \vc_t + \mW_r \vi_t + \mU_r \vs_t^{\prime} \right) \nonumber  \\
	\underline{\vs}_t =& \text{tanh} \left(  \mW \vc_t  + \mW \vi_t  + \vr_t \odot (\mU \vs_t^{\prime} )  \right) \nonumber \\       
	\vs_t =& (1 - \vz_t) \odot \underline{\vs}_t + \vz_t \odot \vs_t^{\prime}
	\end{align}	
	The probabilities for the next target word (from equation \ref{deepout}) also takes into account the new context vector $\vi_t$:
	\begin{equation}
	\mL_o \tanh (\mL_s\vs_t + \mL_c\vc_t +  \mL_i\vi_t + \mL_w\mE_Y[y_{t-1}])
	\end{equation}
	where $\mL_i$ is a new trainable parameter.\\
	In the field of multimodal NMT, the second modality is usually an image computed into feature maps with the help of a CNN.  The annotations $a_1,a_2, \hdots, a_L$ are spatial features  (i.e. each annotation represents features for a specific region in the image) . We follow the same protocol for our experiments and describe it in section \ref{experiment}.
	
	\section{Attention-based Models} \label{ABM}
	
	We evaluate three models of the image attention mechanism $f^\prime_{\text{att}}$ of equation \ref{fatt}. They have in common the fact that at each time step $t$ of the decoding phase, all approaches first take as input the annotation sequence $I$ to derive a time-dependent context vector that contain relevant information in the image to help predict the current target word $y_t$. Even though these models differ in how the time-dependent context vector is derived, they share the same subsequent steps. For each mechanism, we propose two hand-picked illustrations showing where the attention is placed in an image.
	
	\subsection{Soft attention}
	
	Soft attention has firstly been used for syntactic constituency parsing by  \citeauthor{NIPS2015Vinyals} \shortcite{NIPS2015Vinyals}  but has been widely used for translation tasks ever since. One should note that it slightly differs from \citeauthor{BahdanauCB14}  \shortcite{BahdanauCB14} where their attention takes as input  the previous decoder hidden state instead of the current (intermediate) one as shown in equation \ref{fatt}. This mechanism has also been successfully investigated for the task of image description generation \cite{icml2015_xuc15} where a model generates an image's description in natural language. It has been used in multimodal translation as well \cite{CalixtoLC17b}, for which it constitutes a state-of-the-art.\\ \\
	The idea of the soft attentional model is to consider all the annotations when deriving the context vector $\vi_t$.  It consists of a single feed-forward network used to compute an expected alignment $\ve_{t}$ between modality annotation $\va_l$ and the target word to be emitted at the current time step $t$. The inputs are the modality annotations and the intermediate representation of REC1 $\vs_t^{\prime}$:
	
	\begin{equation}
	\ve_{t,l} = \vv^T \tanh(\mU_a \vs_t^{\prime} + \mW_a \va_l ) \label{eqenergy}
	\end{equation}
	The vector $\ve_t$ has length $L$ and its $l$-th item contains a score of how much attention should be put on the $l$-th annotation in order to output the best word at time $t$. We compute normalized scores to create an attention mask $\valpha_t$ over annotations:
	\begin{align}
	\valpha_{t,i} =& \dfrac{\exp(\ve_{t,i})}{\sum\nolimits_{j=1}^L\exp(\ve_{t,j})} \label{eqalpha}\\
	\vi_{t} =& \sum\limits_{i=1}^L \valpha_{t,i}\va_i \label{softsum}
	\end{align}
	Finally, the modality time-dependent context vector $\vi_t$ is computed as a weighted sum over the annotation vectors (equation \ref{softsum}). In the above expressions, $\vv^T$, $\mU_a$ and $\mW_a$ are trained parameters.
	
	\begin{figure}[!ht]
		\centering
		\includegraphics[scale=0.65]{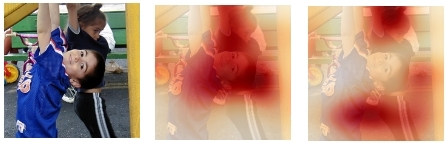}
		\caption{Die beiden \underline{Kinder} spielen auf dem \underline{Spielplatz} .}
	\end{figure}
	
	\begin{figure}[h!]
		\centering
		\includegraphics[scale=0.65]{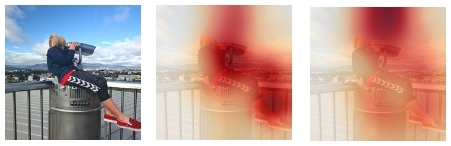}
		\caption{Ein Junge \underline{sitzt} auf und blickt aus einem \underline{Mikroskop} .}
	\end{figure}
	
	\subsection{Hard Stochastic attention} \label{[hard]}
	
	This model is a stochastic and sampling-based process where, at every timestep $t$, we are making a hard choice to attend only one annotation. This corresponds to one spatial location in the image. Hard attention has previously been used in the context of object recognition \cite{NIPS2014_5542,ba-attention-2015} and later extended to image description generation \cite{icml2015_xuc15}. In the context of multimodal NMT, we can follow \citeauthor{icml2015_xuc15} \shortcite{icml2015_xuc15} because both our models involve the same process on images.\\ \\ The mechanism $f_\text{att}$ is now a function that returns a sampled intermediate latent variables $\vgamma_{t,i}$ based upon a multinouilli distribution parameterized by $\valpha$:
	\begin{equation}
	\vgamma_{t} \sim \text{Multinoulli}(\{\valpha_{1, \hdots, L}\}) 
	\end{equation}
	where $\vgamma_{t,i}$ an indicator one-hot variable which is set to 1 if the $i$-th annotation (out of $L$) is the one used to compute the context vector $\vi_t$:
	\begin{align}
	p(\vgamma_{t,l} = 1| \vgamma < t, I) =&  \valpha_{t,l} \label{eq_mc} \\
	\vi_{t} = \sum\limits_{i=1}^L \vgamma_{t,i}\va_i
	\end{align}
	Context vector $\vi_t$ is now seen as the random variable of this distribution. We define the variational
	lower bound $\mathcal{L}(\gamma)$ on the marginal log evidence $\log p(y | I)$ of observing the target sentence $y$ given modality annotations $I$.
	
	\begin{align}
	\mathcal{L}(\gamma) =& \sum\limits_{\gamma} p(\gamma | I) \log p(y|\gamma, I) \nonumber \\
	\leq& \log \sum\limits_{\gamma}p(\gamma | I) p(y|\gamma, I) \nonumber\\
	=& \log p(y|I)	
	\end{align}	
	
	The learning rules can be derived by taking derivatives of the above variational free energy $\mathcal{L}(\gamma)$ with respect to the
	model parameter $\mW$ :
	\begin{align}
	\frac{\partial \mathcal{L}}{\partial \mW} = \sum\limits_{\gamma}p(\gamma|I)& \Bigg[ \frac{\partial \log p(y | \gamma, I)}{\partial \mW} +  \nonumber \\
	&\log p(y | \gamma, I) \frac{\partial \log p(\gamma | I)}{\partial \mW} \Bigg] \label{eq_LW}	
	\end{align}

	In order to propagate a gradient through this process,  the summation in equation \ref{eq_LW} can then be approximated using Monte Carlo based sampling defined by equation \ref{eq_mc}:
	
	\begin{align}
	\frac{\partial \mathcal{L}}{\partial \mW} \approx \frac{1}{N} \sum\limits_{n=1}^{N} & \Bigg[ \frac{\partial \log p(y | \tilde{\gamma}^n, I)}{\partial \mW} +  \nonumber \\
	&\log p(y | \tilde{\gamma}^n, I) \frac{\partial \log p(\tilde{\gamma}^n | I)}{\partial \mW} \Bigg] \label{eq_MC2}
	\end{align}	
	
	To reduce variance of the estimator in equation \ref{eq_MC2}, we use a moving average baseline estimated as an accumulated sum of the previous log likelihoods with exponential decay upon seeing the $k$-th mini-batch:	
	\begin{equation}
	b_k = 0.9 \times b_{k-1} + 0.1 \times \log p(y|\tilde{\gamma}_k, I)
	\end{equation}
	
	\begin{figure}[h!]
		\centering
		\includegraphics[scale=0.60]{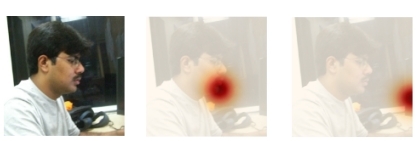}
		\caption{Ein \underline{Mann} sitzt neben einem  \underline{Computerbildschirm} .}
	\end{figure}
	
	\begin{figure}[h!]
		\centering
		\includegraphics[scale=0.65]{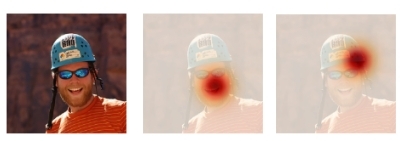}
		\caption{Ein \underline{Mann} in einem orangefarbenen Hemd und mit  \underline{Helm} .}
	\end{figure}
	
	\subsection{Local Attention}
	In this section, we propose a local attentional mechanism that chooses to focus only on a small subset of the image annotations. Local Attention has been used for text-based translation \cite{luongEMNLP} and is inspired by the selective attention model of \citeauthor{gregor15} \shortcite{gregor15} for image generation. Their approach allows the model to select an image patch of varying location and zoom. Local attention uses instead the same "zoom" for all target positions and still achieved good performance. This model can be seen as a trade-off between the soft and hard attentional models. The model picks one patch in the annotation sequence (one spatial location) and selectively focuses on a small window of context around it. Even though an image can't be seen as a temporal sequence, we still hope that the model finds points of interest and selects the useful information around it. This approach has an advantage of being differentiable whereas the stochastic attention requires more complicated techniques such as variance reduction and reinforcement learning to train as shown in section \ref{[hard]}. The soft attention has the drawback to attend the whole image which can be difficult to learn, especially because the number of annotations $L$ is usually large (presumably to keep a significant spatial granularity).
	\\ \\
	More formally, at every decoding step $t$, the model first generates an aligned position $p_t$. Context vector $\vi_t$ is derived as a weighted sum over the annotations within the window $[p_t-D ; p_t+D]$ where $D$ is a fixed model parameter chosen empirically\footnote{We pick $D$ = $|\va_i|/4 = 49$}. These selected annotations correspond to a  squared region in the attention maps around $p_t$. The attention mask $\valpha_t$ is of size $2D+1$. The model predicts $p_t$ as an aligned position in the annotation sequence (referred as Predictive alignment (local-m) in the author's paper) according to the following equation: 	
	\begin{equation}
	p_t = S \cdot \text{sigmoid}(\vv^T\tanh(\mU_a \vs_t^\prime))
	\end{equation}	
	where $\vv^T$ and $\mU_a$ are both trainable model parameters and $S$ is the annotation sequence length $|I|$. Because of the sigmoid, $p_t \in [0, S]$. We use equation \ref{eqenergy} and \ref{eqalpha} respectively to compute the expected alignment vector $\ve_t$ and the attention mask $\valpha_t$. In addition, a Gaussian distribution centered around $p_t$ is placed on the alphas in order to favor annotations near $p_t$:
	
	\begin{equation}
	\valpha_{t,i} = \valpha_{t,i} \exp \bigg(-\frac{(i - p_t)^2}{2\sigma^2}\bigg)
	\end{equation}
	where standard deviation $\sigma = \frac{D}{2}$. We obtain context vector $\vi_t$ by following equation \ref{softsum}.
	\begin{figure}[h!]
		\centering
		\includegraphics[scale=0.60]{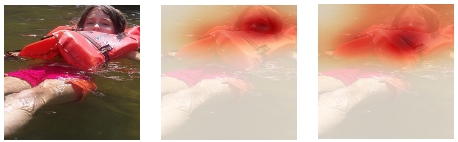}
		\caption{Ein \underline{M{\"a}dchen} mit einer \underline{Schwimmweste} schwimmt im Wasser .}
	\end{figure}
	
	\begin{figure}[h!]
		\centering
		\includegraphics[scale=0.60]{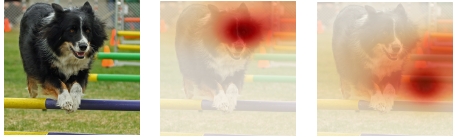}
		\caption{Ein kleiner schwarzer \underline{Hund} springt {\"u}ber \underline{Hindernisse} .}
	\end{figure}

	\section{Image attention optimization} \label{iaopti}
	Three optimizations can be added to the attention mechanism regarding the image modality. All lead to a better use of the image by the model and improved the translation scores overall.\\ \\
	At every decoding step $t$, we compute a gating scalar $\vbeta_t \in [0, 1]$ according to the previous decoder state $\vs_{t-1}$:    
	\begin{equation}
	\vbeta_t = \sigma(\mW_\beta \vs_{t-1} + \vb_\beta)
	\end{equation}
	It is then used to compute the time-dependent image context vector :    
	\begin{equation}
	\vi_t = \vbeta_t \sum\limits_{l=1}^L \valpha_{t,l} \va_l
	\end{equation}    
	\citeauthor{icml2015_xuc15} \shortcite{icml2015_xuc15} empirically found it to put more emphasis on the objects in the image descriptions  generated with their model.\\ \\	
	We also double the output size of trainable parameters $\mU_a$, $\mW_a$ and $\vv^T$ in equation \ref{eqenergy} when it comes to compute the expected annotations over the image annotation sequence. More formally, given the image annotation sequence $I=(\va_1,\va_2, \hdots, \va_L), \va_i \in \mathbb{R}^D$, the tree matrices are of size $D\times2D$, $D\times2D$ and $2D\times1$ respectively. We noticed a better coverage of the objects in the image by the alpha weights.\\ \\
	Lastly, we use a grounding attention inspired by  \citeauthor{delbrouck2017multimodal}  \shortcite{delbrouck2017multimodal}. The mechanism merge each spatial location $\va_i$ in the annotation sequence $I$ with the initial decoder state $\vs_0$ obtained in equation \ref{initdec} with non-linearity :
	\begin{align}
	I^{\prime} =& (f(\va_1 + \vs_0), f(\va_2 + \vs_0), \hdots, f(\va_L + \vs_0))
	\end{align}
	where $f$ is $\tanh$ function. The new annotations go through a L2 normalization layer followed by two $1 \times 1$ convolutional layers (of size $D \rightarrow 512, 512 \rightarrow 1$ respectively) to obtain $L \times 1$ weights, one for each spatial location. We normalize the weights with a softmax to obtain a soft attention map $\valpha$. Each annotation $\va_i$ is then weighted according to its corresponding $\valpha_i$:
	\begin{align}
	I =& (\valpha_1\va_1, \valpha_2\va_2, \hdots, \valpha_L\va_L)
	\end{align}
	This method can be seen as the removal of unnecessary information in the image annotations according to the source sentence. This attention is used on top of the others - before decoding - and is referred as "grounded image" in Table \ref{score-tabular}.

	\begin{table*}
		\centering
		\begin{tabular}{lcccccc}
			\multicolumn{1}{c}{\bf Model}  &\multicolumn{6}{c}{\bf 				Test Scores}
			\\ \hline \\
			&BLEU$\uparrow$& & METEOR$\uparrow$& &TER$\downarrow$ & \\
			\textbf{Monomodal (text only)} \\
			\citeauthor{caglayan2016does} \shortcite{caglayan2016does}            &32.50& &49.2& & & \\
			\citeauthor{CalixtoLC17b} \shortcite{CalixtoLC17b}            &33.70& & 52.3& & 46.7&  \\
			NMT         &34.11&\textcolor{olive}{$\uparrow$ +0.41} &52.4&\textcolor{olive}{$\uparrow$ +0.1} &46.2&\textcolor{olive}{$\downarrow$ -0.5}  \\
			\hline \\
			\textbf{Multimodal} \\
			\citeauthor{caglayan2016does} \shortcite{caglayan2016does}    &27.82& &45.0& &-& \\
			\citeauthor{huang2016attention} \shortcite{huang2016attention}    &36.50& &54.1& &- \\          
			\citeauthor{CalixtoLC17b} \shortcite{CalixtoLC17b}
			&36.50& &55.0& &43.7&  \\
			Soft attention &37.10 & \textcolor{olive}{$\uparrow$ +0.60}&54.8 & \textcolor{red}{$\downarrow$ -0.2}&42.8& \textcolor{olive}{$\downarrow$ -0.9}  \\
			Local attention &37.55 & \textcolor{olive}{$\uparrow$ +1.05}&54.8 & \textcolor{red}{$\downarrow$ -0.2}&42.4& \textcolor{olive}{$\downarrow$ -1.3}  \\
			Stochastic attention &38.01& \textcolor{olive}{$\uparrow$ +1.51}&55.4 &\textcolor{olive}{$\uparrow$ +0.4} &41.5& \textcolor{olive}{$\downarrow$ -2.2} \\        
			Soft attention + grounded image&37.62& \textcolor{olive}{$\uparrow$ +1.12} &55.3&  \textcolor{olive}{$\uparrow$ +0.3} &41.8 & \textcolor{olive}{$\downarrow$ -1.9} \\
			Stochastic attention + grounded image&38.17&\textcolor{olive}{$\uparrow$ +1.67} &55.4 &\textcolor{olive}{$\uparrow$ +0.4} &41.5& \textcolor{olive}{$\downarrow$ -2.2}      
		\end{tabular}      
		\caption{Results on the 1000 test triples of the Multi30K dataset. We pick \citeauthor{CalixtoLC17b} \shortcite{CalixtoLC17b} scores as baseline and report our results accordingly (green for improvement and red for deterioration). In each of our experiments, Soft attention is used for text. The comparison is hence with respect to the attention mechanism used for the image modality.}
        \label{score-tabular}
	\end{table*}%

	\section{Experiments} \label{experiment}
	For this experiments on Multimodal Machine Translation, we used the Multi30K dataset \cite{elliott-EtAl:2016:VL16} which is an extended version of the Flickr30K Entities. For each image, one of the English descriptions was selected and manually translated into German by a professional translator. As training and development data, 29,000 and 1,014 triples are used respectively. A test set of size 1000 is used for metrics evaluation.

	\subsection{Training and model details}
	
	All our models are build on top of the nematus framework \cite{nematus}. The encoder is a bidirectional RNN with GRU, one 1024D single-layer forward and one 1024D single-layer backward RNN. Word embeddings for source and target language are of 620D and trained jointly with the model. Word embeddings and other non-recurrent matrices are initialized by sampling from a Gaussian $\mathcal{N}(0, 0.01^2)$, recurrent matrices are random orthogonal and bias vectors are all initialized to zero. \\ \\   
	To create the image annotations used by our decoder, we used a ResNet-50 pre-trained on ImageNet and extracted the features of size $14 \times 14 \times 1024$ at its res4f layer \cite{He_2016_CVPR}. In our experiments, our decoder operates on the flattened 196 $\times$ 1024 (i.e $L \times D$).
	We also apply dropout with a probability of 0.5 on the embeddings, on the hidden states in the bidirectional RNN in the encoder as well as in the decoder. In the decoder, we also apply dropout on the text annotations $\vh_i$, the image features $\va_i$, on both modality context vector and on all components of the deep output layer before the readout operation. We apply dropout using one same mask in all time steps \cite{Gal2016Theoretically}. \\ \\    
	We also normalize and tokenize English and German descriptions using the Moses tokenizer scripts \cite{Koehn:2007}. We use the byte pair encoding algorithm on the train set to convert space-separated tokens into subwords \citep{sennrich2016subword}, reducing our vocabulary size to 9226 and 14957 words for English and German respectively.\\ \\
	All variants of our attention model were trained with ADADELTA \cite{ADADELTAZeiler}, with mini-batches of size 80 for our monomodal (text-only) NMT model and 40 for our multimodal NMT. We apply early stopping for model selection based on BLEU4 : training is halted if no improvement on the development set is observed for more than 20 epochs. We use the metrics BLEU4 \cite{Papineni:2002}, METEOR \cite{meteor-wmt:2014} and TER \cite{Snover06astudy} to evaluate the quality of our models' translations.
	
	\subsection{Quantitative results}
	We notice a nice overall progress over \citeauthor{CalixtoLC17b} \shortcite{CalixtoLC17b} multimodal baseline, especially when using the stochastic attention. With improvements of +1.51 BLEU and -2.2 TER on both precision-oriented metrics, the model shows a strong similarity of the n-grams of our candidate translations with respect to the references. The more recall-oriented metrics METEOR scores are roughly the same across our models which is expected because all attention mechanisms share the same subsequent step at every time-step $t$, i.e. taking into account the attention weights of previous time-step $t-1$ in order to compute the new intermediate hidden state proposal and therefore the new context vector $i_t$. Again, the largest improvement is given by the hard stochastic attention mechanism (+0.4 METEOR): because it is modeled as a decision process according to the previous choices, this may reinforce the idea of recall. We also remark interesting improvements when using the grounded mechanism, especially for the soft attention. The soft attention may benefit more of the grounded image because of the wide range of spatial locations it looks at, especially compared to the stochastic attention. This motivates us to dig into more complex grounding techniques in order to give the machine a deeper understanding of the modalities. \\ \\
	Note that even though our baseline NMT model is basically the same as \citeauthor{CalixtoLC17b} \shortcite{CalixtoLC17b}, our experiments results are slightly better. This is probably due to the different use of dropout and subwords. We also compared our results to  \citeauthor{caglayan2016does} \shortcite{caglayan2016does} because our multimodal models are nearly identical with the major exception of the gating scalar (cfr. section \ref{iaopti}). This motivated some of our qualitative analysis and hesitation towards the current architecture in the next section.
	
	\begin{figure*} 
		\includegraphics[scale=0.45]{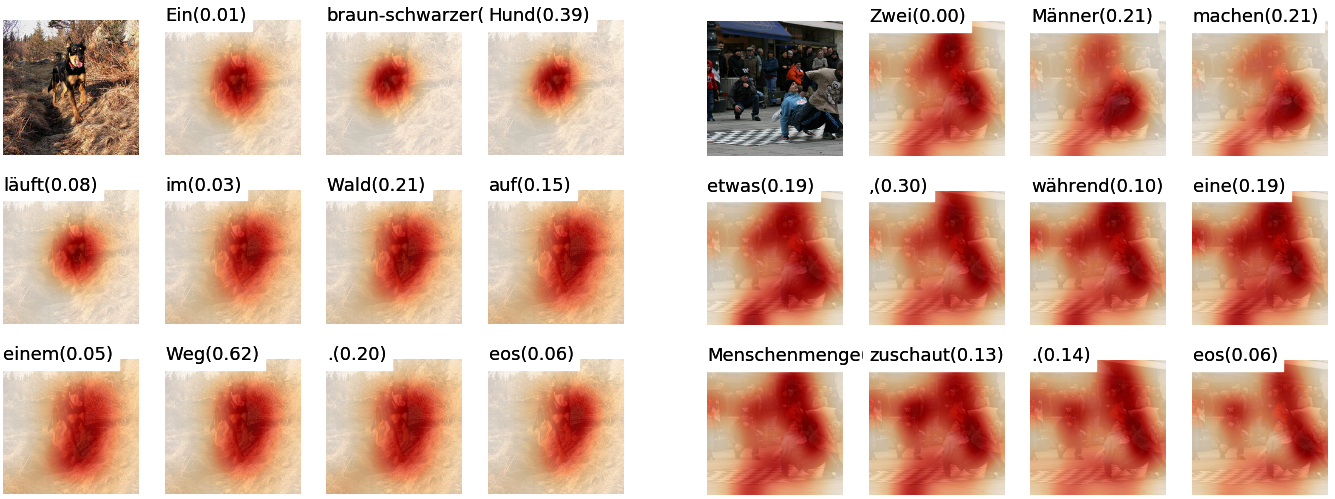}
		\caption{Representative figures of the soft-attention behavior discussed in \S\ref{qualit}}
		\label{anal1}
	\end{figure*}

	\subsection{Qualitative results} \label{qualit}
\begin{figure}[h] 
		\centering
		\includegraphics[scale=0.55]{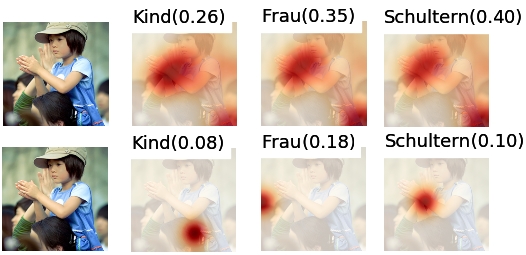} 
        \caption{Wrong detection for both Soft attention (top) and Hard stochastic attention (bottom)}
        \label{last}
	\end{figure}

	For space-saving and ergonomic reasons, we only discuss about the hard stochastic and soft attention, the latter being a generalization of the local attention. \\
	As we can see in Figure \ref{anal1}, the soft attention model is looking roughly at the same region of the image for every decoding step $t$. Because the words "\textit{hund}"(dog), "\textit{wald}"(forest) or "\textit{weg}"(way) in left image are objects, they benefit from a high gating scalar. As a matter of fact, the attention mechanism has learned to detect the objects within a scene (at every time-step, whichever word we are decoding as shown in the right image) and the gating scalar has learned to decide whether or not we have to look at the picture (or more accurately whether or not we are translating an object). Without this scalar, the translation scores undergo a massive drop (as seen in \citeauthor{caglayan2016does} \shortcite{caglayan2016does}) which means that the attention mechanisms don't really understand the more complex relationships between objects, what is really happening in the scene. Surprisingly, the gating scalar happens to be really low in the stochastic attention mechanism: a significant amount of sentences don't have a summed gating scalar $\geq$ 0.10. The model totally discards the image in the translation process. \\ \\
	It is also worth to mention that we use a ResNet trained on 1.28 million images for a classification tasks. The features used by the attention mechanism are strongly object-oriented and the machine could miss important information for a multimodal translation task. We believe that the robust architecture of both encoders $\{\overleftarrow{\Psi}_\text{enc},\overrightarrow{\Psi}_\text{enc}\}$ combined with a GRU layer and word-embeddings took care of the right translation for relationships between objects and time-dependencies. Yet, we noticed a common misbehavior for all our multimodal models: if the attention loose track of the objects in the picture and "gets lost", the model still takes it into account and somehow overrides the information brought by the text-based annotations. The translation is then totally mislead. We illustrate with an example:
	\begin{table}[h!]
		\centering
		\begin{tabular}{|ll|}
			\hline
			Ref:&Ein Kind sitzt auf den Schultern einer \\
			&Frau und klatscht .\\
			Mono:&Ein Kind sitzt auf den Schultern einer \\
			&Frau und schl{\"a}ft .\\
			Soft:&Ein Kind , das sich auf der Schultern \\
			&eines Frau reitet , f{\"a}hrt auf den\\ 
			&Schultern .\\
			Hard:&Ein Kind in der Haltung , w{\"a}hrend er \\
			&auf den Schultern einer Frau f{\"a}hrt .\\
			\hline
		\end{tabular}
	\end{table}\\
    The monomodal translation has a sentence-level BLEU of 82.16 whilst the soft attention and hard stochastic attention scores are of 16.82 and 34.45 respectively. Figure \ref{last} shows the attention maps for both mechanism. Nevertheless, one has to concede that the use of images indubitably helps the translation as shown in the score tabular.

	\section{Conclusion and future work} \label{concl}
	
	We have tried different attention mechanism and tweaks for the image modality. We showed improvements and encouraging results overall on the Flickr30K Entities dataset. Even though we identified some flaws of the current attention mechanisms, we can conclude pretty safely that images are an helpful resource for the machine in a translation task. We are looking forward to try out richer and more suitable features for multimodal translation (ie. dense captioning features). Another interesting approach would be to use visually grounded word embeddings to capture visual notions of semantic relatedness.

\section{Acknowledgements}
This work was partly supported by the Chist-Era project IGLU with contribution from the Belgian Fonds de la Recherche Scientique (FNRS), contract no. R.50.11.15.F, and by the FSO project VCYCLE with contribution from the Belgian Waloon Region, contract no. 1510501.

	\bibliography{emnlp2017}

\begin{thebibliography}{}
\expandafter\ifx\csname natexlab\endcsname\relax\def\natexlab#1{#1}\fi

\bibitem[{Ba et~al.(2015)Ba, Mnih, and Kavukcuoglu}]{ba-attention-2015}
Jimmy Ba, Volodymyr Mnih, and Koray Kavukcuoglu. 2015.
\newblock Multiple object recognition with visual attention.
\newblock In {\em Proceedings of the International Conference on Learning
  Representations (ICLR)\/}.

\bibitem[{Bahdanau et~al.(2014)Bahdanau, Cho, and Bengio}]{BahdanauCB14}
Dzmitry Bahdanau, Kyunghyun Cho, and Yoshua Bengio. 2014.
\newblock Neural machine translation by jointly learning to align and
  translate.
\newblock {\em CoRR\/} abs/1409.0473.

\bibitem[{Caglayan et~al.(2016)Caglayan, Aransa, Wang, Masana,
  Garc\'{i}a-Mart\'{i}nez, Bougares, Barrault, and van~de
  Weijer}]{caglayan2016does}
Ozan Caglayan, Walid Aransa, Yaxing Wang, Marc Masana, Mercedes
  Garc\'{i}a-Mart\'{i}nez, Fethi Bougares, Lo\"{i}c Barrault, and Joost van~de
  Weijer. 2016.
\newblock \href{http://www.aclweb.org/anthology/W/W16/W16-2358}{Does
  multimodality help human and machine for translation and image captioning?}
\newblock In {\em Proceedings of the First Conference on Machine
  Translation\/}. Association for Computational Linguistics, Berlin, Germany,
  pages 627--633.
\newblock
  \href{http://www.aclweb.org/anthology/W/W16/W16-2358}{http://www.aclweb.org/anthology/W/W16/W16-2358}.

\bibitem[{Calixto et~al.(2017)Calixto, Liu, and Campbell}]{CalixtoLC17b}
Iacer Calixto, Qun Liu, and Nick Campbell. 2017.
\newblock \href{http://arxiv.org/abs/1702.01287}{Doubly-attentive decoder for
  multi-modal neural machine translation}.
\newblock {\em CoRR\/} abs/1702.01287.
\newblock
  \href{http://arxiv.org/abs/1702.01287}{http://arxiv.org/abs/1702.01287}.

\bibitem[{Cho et~al.(2014)Cho, van Merri{\"{e}}nboer, G{\"{u}}l{\c c}ehre,
  Bahdanau, Bougares, Schwenk, and Bengio}]{cho-al-emnlp14}
Kyunghyun Cho, Bart van Merri{\"{e}}nboer, {\c C}ağlar G{\"{u}}l{\c c}ehre,
  Dzmitry Bahdanau, Fethi Bougares, Holger Schwenk, and Yoshua Bengio. 2014.
\newblock \href{http://www.aclweb.org/anthology/D14-1179}{Learning phrase
  representations using rnn encoder--decoder for statistical machine
  translation}.
\newblock In {\em Proceedings of the 2014 Conference on Empirical Methods in
  Natural Language Processing (EMNLP)\/}. Association for Computational
  Linguistics, Doha, Qatar, pages 1724--1734.
\newblock
  \href{http://www.aclweb.org/anthology/D14-1179}{http://www.aclweb.org/anthology/D14-1179}.

\bibitem[{Chung et~al.(2014)Chung, Gulcehre, Cho, and Bengio}]{ChungGCB14}
Junyoung Chung, Caglar Gulcehre, Kyunghyun Cho, and Yoshua Bengio. 2014.
\newblock {\em Empirical evaluation of gated recurrent neural networks on
  sequence modeling\/}.

\bibitem[{Delbrouck and Dupont(2017)}]{delbrouck2017multimodal}
Jean-Benoit Delbrouck and Stephane Dupont. 2017.
\newblock \href{https://arxiv.org/pdf/1703.08084.pdf}{Multimodal compact
  bilinear pooling for multimodal neural machine translation}.
\newblock {\em arXiv preprint arXiv:1703.08084\/}
  \href{https://arxiv.org/pdf/1703.08084.pdf}{https://arxiv.org/pdf/1703.08084.pdf}.

\bibitem[{Denkowski and Lavie(2014)}]{meteor-wmt:2014}
Michael Denkowski and Alon Lavie. 2014.
\newblock Meteor universal: Language specific translation evaluation for any
  target language.
\newblock In {\em Proceedings of the EACL 2014 Workshop on Statistical Machine
  Translation\/}.

\bibitem[{{Elliott} et~al.(2016){Elliott}, {Frank}, {Sima'an}, and
  {Specia}}]{elliott-EtAl:2016:VL16}
D.~{Elliott}, S.~{Frank}, K.~{Sima'an}, and L.~{Specia}. 2016.
\newblock Multi30k: Multilingual english-german image descriptions pages
  70--74.

\bibitem[{Gal and Ghahramani(2016)}]{Gal2016Theoretically}
Yarin Gal and Zoubin Ghahramani. 2016.
\newblock A theoretically grounded application of dropout in recurrent neural
  networks.
\newblock In {\em Advances in Neural Information Processing Systems 29
  (NIPS)\/}.

\bibitem[{Gregor et~al.(2015)Gregor, Danihelka, Graves, Rezende, and
  Wierstra}]{gregor15}
Karol Gregor, Ivo Danihelka, Alex Graves, Danilo Rezende, and Daan Wierstra.
  2015.
\newblock \href{http://proceedings.mlr.press/v37/gregor15.html}{Draw: A
  recurrent neural network for image generation}.
\newblock In Francis Bach and David Blei, editors, {\em Proceedings of the 32nd
  International Conference on Machine Learning\/}. PMLR, Lille, France,
  volume~37 of {\em Proceedings of Machine Learning Research\/}, pages
  1462--1471.
\newblock
  \href{http://proceedings.mlr.press/v37/gregor15.html}{http://proceedings.mlr.press/v37/gregor15.html}.

\bibitem[{He et~al.(2016)He, Zhang, Ren, and Sun}]{He_2016_CVPR}
Kaiming He, Xiangyu Zhang, Shaoqing Ren, and Jian Sun. 2016.
\newblock Deep residual learning for image recognition.
\newblock In {\em The IEEE Conference on Computer Vision and Pattern
  Recognition (CVPR)\/}.

\bibitem[{Huang et~al.(2016)Huang, Liu, Shiang, Oh, and
  Dyer}]{huang2016attention}
Po-Yao Huang, Frederick Liu, Sz-Rung Shiang, Jean Oh, and Chris Dyer. 2016.
\newblock Attention-based multimodal neural machine translation.
\newblock In {\em Proceedings of the First Conference on Machine Translation,
  Berlin, Germany\/}.

\bibitem[{Koehn et~al.(2007)Koehn, Hoang, Birch, Callison-Burch, Federico,
  Bertoldi, Cowan, Shen, Moran, Zens, Dyer, Bojar, Constantin, and
  Herbst}]{Koehn:2007}
Philipp Koehn, Hieu Hoang, Alexandra Birch, Chris Callison-Burch, Marcello
  Federico, Nicola Bertoldi, Brooke Cowan, Wade Shen, Christine Moran, Richard
  Zens, Chris Dyer, Ond\v{r}ej Bojar, Alexandra Constantin, and Evan Herbst.
  2007.
\newblock \href{http://dl.acm.org/citation.cfm?id=1557769.1557821}{Moses: Open
  source toolkit for statistical machine translation}.
\newblock In {\em Proceedings of the 45th Annual Meeting of the ACL on
  Interactive Poster and Demonstration Sessions\/}. Association for
  Computational Linguistics, Stroudsburg, PA, USA, ACL '07, pages 177--180.
\newblock
  \href{http://dl.acm.org/citation.cfm?id=1557769.1557821}{http://dl.acm.org/citation.cfm?id=1557769.1557821}.

\bibitem[{Luong et~al.(2015)Luong, Pham, and Manning}]{luongEMNLP}
Minh-Thang Luong, Hieu Pham, and Christopher~D. Manning. 2015.
\newblock Effective approaches to attention-based neural machine translation.
\newblock In {\em Proceedings of the 2015 Conference on Empirical Methods in
  Natural Language Processing\/}.

\bibitem[{Mnih et~al.(2014)Mnih, Heess, Graves, and
  kavukcuoglu}]{NIPS2014_5542}
Volodymyr Mnih, Nicolas Heess, Alex Graves, and koray kavukcuoglu. 2014.
\newblock
  \href{http://papers.nips.cc/paper/5542-recurrent-models-of-visual-attention.pdf}{Recurrent
  models of visual attention}.
\newblock In Z.~Ghahramani, M.~Welling, C.~Cortes, N.D. Lawrence, and K.Q.
  Weinberger, editors, {\em Advances in Neural Information Processing Systems
  27\/}, Curran Associates, Inc., pages 2204--2212.
\newblock
  \href{http://papers.nips.cc/paper/5542-recurrent-models-of-visual-attention.pdf}{http://papers.nips.cc/paper/5542-recurrent-models-of-visual-attention.pdf}.

\bibitem[{Papineni et~al.(2002)Papineni, Roukos, Ward, and Zhu}]{Papineni:2002}
Kishore Papineni, Salim Roukos, Todd Ward, and Wei-Jing Zhu. 2002.
\newblock \href{https://doi.org/10.3115/1073083.1073135}{Bleu: A method for
  automatic evaluation of machine translation}.
\newblock In {\em Proceedings of the 40th Annual Meeting on Association for
  Computational Linguistics\/}. Association for Computational Linguistics,
  Stroudsburg, PA, USA, ACL '02, pages 311--318.
\newblock
  \href{https://doi.org/10.3115/1073083.1073135}{https://doi.org/10.3115/1073083.1073135}.

\bibitem[{Pascanu et~al.(2014)Pascanu, Gulcehre, Cho, and Bengio}]{Pascanu2014}
Razvan Pascanu, Caglar Gulcehre, Kyunghyun Cho, and Yoshua Bengio. 2014.
\newblock {\em How to construct deep recurrent neural networks\/}.

\bibitem[{Sennrich et~al.(2017)Sennrich, Firat, Cho, Birch, Haddow, Hitschler,
  Junczys-Dowmunt, L{"a}ubli, {Miceli Barone}, Mokry, and Nadejde}]{nematus}
Rico Sennrich, Orhan Firat, Kyunghyun Cho, Alexandra Birch, Barry Haddow,
  Julian Hitschler, Marcin Junczys-Dowmunt, Samuel L{"a}ubli, Antonio~Valerio
  {Miceli Barone}, Jozef Mokry, and Maria Nadejde. 2017.
\newblock {Nematus: a Toolkit for Neural Machine Translation}.
\newblock In {\em {Proceedings of the Demonstrations at the 15th Conference of
  the European Chapter of the Association for Computational Linguistics}\/}.

\bibitem[{Sennrich et~al.(2016)Sennrich, Haddow, and
  Birch}]{sennrich2016subword}
Rico Sennrich, Barry Haddow, and Alexandra Birch. 2016.
\newblock \href{http://www.aclweb.org/anthology/P16-1162}{{Neural Machine
  Translation of Rare Words with Subword Units}}.
\newblock In {\em In Proceedings of the 54th Annual Meeting of the Association
  for Computational Linguistics (Volume 1: Long Papers)\/}.
\newblock
  \href{http://www.aclweb.org/anthology/P16-1162}{http://www.aclweb.org/anthology/P16-1162}.

\bibitem[{Snover et~al.(2006)Snover, Dorr, Schwartz, Micciulla, and
  Makhoul}]{Snover06astudy}
Matthew Snover, Bonnie Dorr, Richard Schwartz, Linnea Micciulla, and John
  Makhoul. 2006.
\newblock A study of translation edit rate with targeted human annotation.
\newblock In {\em In Proceedings of Association for Machine Translation in the
  Americas\/}. pages 223--231.

\bibitem[{Specia et~al.(2016)Specia, Frank, Sima'an, and
  Elliott}]{specia-EtAl:2016:WMT}
Lucia Specia, Stella Frank, Khalil Sima'an, and Desmond Elliott. 2016.
\newblock \href{http://www.aclweb.org/anthology/W/W16/W16-2346}{A shared task
  on multimodal machine translation and crosslingual image description}.
\newblock In {\em Proceedings of the First Conference on Machine
  Translation\/}. Association for Computational Linguistics, Berlin, Germany,
  pages 543--553.
\newblock
  \href{http://www.aclweb.org/anthology/W/W16/W16-2346}{http://www.aclweb.org/anthology/W/W16/W16-2346}.

\bibitem[{Sutskever et~al.(2014)Sutskever, Vinyals, and Le}]{SutskeverVL14}
Ilya Sutskever, Oriol Vinyals, and Quoc~V Le. 2014.
\newblock Sequence to sequence learning with neural networks.
\newblock In {\em Advances in neural information processing systems\/}. pages
  3104--3112.

\bibitem[{Vinyals et~al.(2015)Vinyals, Kaiser, Koo, Petrov, Sutskever, and
  Hinton}]{NIPS2015Vinyals}
Oriol Vinyals, \L~ukasz Kaiser, Terry Koo, Slav Petrov, Ilya Sutskever, and
  Geoffrey Hinton. 2015.
\newblock
  \href{http://papers.nips.cc/paper/5635-grammar-as-a-foreign-language.pdf}{Grammar
  as a foreign language}.
\newblock In C.~Cortes, N.~D. Lawrence, D.~D. Lee, M.~Sugiyama, and R.~Garnett,
  editors, {\em Advances in Neural Information Processing Systems 28\/}, Curran
  Associates, Inc., pages 2773--2781.
\newblock
  \href{http://papers.nips.cc/paper/5635-grammar-as-a-foreign-language.pdf}{http://papers.nips.cc/paper/5635-grammar-as-a-foreign-language.pdf}.

\bibitem[{Xu et~al.(2015)Xu, Ba, Kiros, Cho, Courville, Salakhudinov, Zemel,
  and Bengio}]{icml2015_xuc15}
Kelvin Xu, Jimmy Ba, Ryan Kiros, Kyunghyun Cho, Aaron Courville, Ruslan
  Salakhudinov, Rich Zemel, and Yoshua Bengio. 2015.
\newblock \href{http://jmlr.org/proceedings/papers/v37/xuc15.pdf}{Show, attend
  and tell: Neural image caption generation with visual attention}.
\newblock In David Blei and Francis Bach, editors, {\em Proceedings of the 32nd
  International Conference on Machine Learning (ICML-15)\/}. JMLR Workshop and
  Conference Proceedings, pages 2048--2057.
\newblock
  \href{http://jmlr.org/proceedings/papers/v37/xuc15.pdf}{http://jmlr.org/proceedings/papers/v37/xuc15.pdf}.

\bibitem[{Zeiler(2012)}]{ADADELTAZeiler}
Matthew~D. Zeiler. 2012.
\newblock \href{http://arxiv.org/abs/1212.5701}{{ADADELTA:} an adaptive
  learning rate method}.
\newblock {\em CoRR\/} abs/1212.5701.
\newblock
  \href{http://arxiv.org/abs/1212.5701}{http://arxiv.org/abs/1212.5701}.

\end{thebibliography}
	\bibliographystyle{emnlp_natbib}
	
\end{document}